\documentclass{sig-alternate}

\usepackage{url}

\begin{document}

\permission{Permission to make digital or hard copies of all or part of this work for personal or classroom use is granted without fee provided that copies are not made or distributed for profit or commercial advantage and that copies bear this notice and the full citation on the first page. Copyrights for components of this work owned by others than ACM must be honored. Abstracting with credit is permitted. To copy otherwise, or republish, to post on servers or to redistribute to lists, requires prior specific permission and/or a fee. Request permissions from Permissions@acm.org.}
\conferenceinfo{K-CAP 2015,}{October 07-10, 2015, Palisades, NY, USA}
\copyrightetc{\copyright~2015 ACM. ISBN \the\acmcopyr}
\crdata{978-1-4503-3849-3/15/10\$15.00\\
DOI: http://dx.doi.org/10.1145/2815833.2815841}

\title{Improving the Competency of First-Order Ontologies}

\numberofauthors{3}

\author{
\alignauthor
Javier \'Alvez\\
       \affaddr{LoRea Group}\\
       \affaddr{University of the Basque Country UPV/EHU}\\
       \email{javier.alvez@ehu.eus}
\alignauthor
Paqui Lucio\\
       \affaddr{LoRea Group}\\
       \affaddr{University of the Basque Country UPV/EHU}\\
       \email{paqui.lucio@ehu.eus}
\alignauthor German Rigau\\
       \affaddr{IXA NLP Group}\\
       \affaddr{University of the Basque Country UPV/EHU}\\
       \email{german.rigau@ehu.eus}
}

\date{\today}

\maketitle


\newcommand{\tab}{\hspace{0.2cm}}
\newcommand{\connective}[1]{\bf #1 \;}
\newcommand{\predicate}[1]{\rm #1}
\newcommand{\constant}[1]{\rm #1}
\newcommand{\variable}[1]{\tt ?#1}

\newcommand{\textConstant}[1]{{\it{#1}}}
\newcommand{\textPredicate}[1]{{\it{#1}}}

\begin{abstract}

We introduce a new framework to evaluate and improve first-order (FO) ontologies using automated theorem provers (ATPs) on the basis of competency questions (CQs). Our framework includes both the adaptation of a methodology for evaluating ontologies to the framework of first-order logic and a new set of non-trivial CQs designed to evaluate FO versions of SUMO, which significantly extends the very small set of CQs proposed in the literature. Most of these new CQs have been automatically generated from a small set of patterns and the mapping of WordNet to SUMO. Applying our framework, we demonstrate that Adimen-SUMO v2.2 outperforms TPTP-SUMO. In addition, using the feedback provided by ATPs we have set an improved version of Adimen-SUMO (v2.4). This new version outperforms the previous ones in terms of competency. For instance, ``{\it Humans can reason}'' is automatically inferred from Adimen-SUMO v2.4, while it is neither deducible from TPTP-SUMO nor Adimen-SUMO v2.2.

\end{abstract}



\section{Introduction}

Ontologies are being used in a wide range of applications and knowledge based systems \cite{AnV08}. Like any other component of a system, an ontology requires a repetitive process of refinement and evaluation during its development and application lifecycle. Ontologies can be evaluated by considering their use in an application when performing correct predictions on inferencing \cite{PoM04}. In order to enable better reasoning capabilities, the inferencing process should be able to deduce from the ontology as much correct implicit knowledge as possible. In \cite{GrF95}, the authors propose to use a set of competency questions (CQs) to evaluate an ontology, which are goals that the ontology is expected to answer. The proposed methodology can be applied to any formal ontology if there exists a decision algorithm for the underlying logic.

In general the process of obtaining CQs is not automatic but creative \cite{FGS13}. Depending on the size and complexity of the ontology, the process of creating a suitable set of CQs is by itself a very challenging and costly task.
 
Although OWL-DL \cite{HoP04} is currently one of the most common formal knowledge representation formalisms, it is unable to cope with expressive ontologies like Cyc \cite{Matuszek+'06}, DOLCE \cite{Gangemi+'02} or SUMO \cite{Niles+Pease'01}. Fortunately, state-of-the-art automatic theorem provers (ATPs) for first-order logic (FOL) like Vampire \cite{RiV02} or E \cite{Sch02} are highly sophisticated systems that have been proved to provide advanced reasoning support to substantial FOL conversions of expressive ontologies, including first-orderized Cyc \cite{RRG05}, SUMO \cite{HoV06}, TPTP-SUMO \cite{PeS07} and Adimen-SUMO \cite{ALR12}. Despite these preliminary experiments, as far as we know, there is no previous work that applies the methodology of \cite{GrF95} to first-order (FO) ontologies using FOL ATPs. 

\begin{table*}[t]
\caption{\label{table:SUMOFigures} Some figures about SUMO, TPTP-SUMO and Adimen-SUMO}
\centering
\begin{tabular} {lrrr}
\hline \\[-6pt]
 & \hspace{0.5cm} {\bf SUMO} & \hspace{0.5cm} {\bf TPTP-SUMO} & \hspace{0.5cm} {\bf Adimen-SUMO} \\
\hline\rule{-4pt}{10pt}
Objects & 20,081 & 2,920 & 1,009 \\
Classes &  5,563 & 2,086 & 2,124 \\
Relations &  369 & 208 &  208 \\
Attributes &  2,153 & 68 & 66 \\
{\bf Total} & {\bf 28,166} & {\bf 5,282} & {\bf 3,407} \\
\hline
\end{tabular}
\end{table*}

The contributions of this paper are manyfold. First, following \cite{GrF95}, we present a new framework to evaluate and improve FO ontologies using ATPs. Second, we introduce a new set of very large and non-trivial CQs designed to evaluate FO versions of SUMO. Our set includes 64 creative CQs for development and more than 7,000 automatically generated CQs for testing. Our set of creative CQs extends the 33 questions from the CSR (Common Sense Reasoning) problem domain of TPTP (Thousands of Problems for Theorem Provers) \cite{Sut09} and the 5 questions described in \cite{ALR12}. Additionally, we have also devised and developed a novel automatic procedure for generating a very large set of non-trivial CQs from a small set of conceptual patterns on the basis of the knowledge encoded in WordNet (WN) \cite{Fellbaum'98} and its mapping to SUMO \cite{Niles+Pease'03}. Third, as a result of the application of our framework, we create a new version of Adimen-SUMO. Finally, using our new framework and CQs, we carry out an empirical comparison of the existing FO versions of SUMO. According to our experimental results, the new version of Adimen-SUMO outperforms TPTP-SUMO and all previous versions of Adimen-SUMO. For example, from Adimen-SUMO v2.4 ATPs infer that ``{\it Tables do not have a brain}'' and ``{\it Humans can reason}'',  but ``{\it Organisms cannot be dead}'' and ``{\it Tables can eat}'' are not inferred. However, ATPs yield the opposite results from TPTP-SUMO and Adimen-SUMO v2.2 in the four cases. Both the new version of Adimen-SUMO and the new set of CQs are freely available.\footnote{ \url{http://adimen.si.ehu.es/web/AdimenSUMO}}

Obviously, this type of non-trivial inferences could be very useful for a wide range of knowledge intensive applications. For instance, to help validating the consistency of associated semantic resources like WN, or to derive new explicit knowledge from them. Furthermore, WN is being used world-wide to anchor different types of semantic resources and wordnets in many languages.\footnote{\url{http://www.globalwordnet.org/}} Therefore, similar inferences can be obtained for other semantic resources and languages other than English. Likewise, WN is connected to several databases such as OpenCyc \cite{reed2002}, DBpedia \cite{auer2007,bizer2009} or YAGO \cite{Hoffart+'12} thanks to the Linked Open Data (LOD) cloud initiative \cite{Bizer+'09}. The interlinking of these diverse databases to a fully operational upper level ontology promises a ``Web of Data'' that will enable machines to more easily exploit its content \cite{jain2010}.

In the next two sections, we first introduce SUMO and its FOL versions, and then our adaptation of the methodology proposed by \cite{GrF95}. In Section \ref{section:refining}, we illustrate the process of improving an ontology, which yields Adimen-SUMO v2.4, by providing some examples.  Next, the process of automatically obtaining a new set of CQs from WN and its mapping to SUMO is described in Section \ref{section:benchmark}. In Section \ref{section:evaluation}, we report on the competency of TPTP-SUMO and the different versions of Adimen-SUMO. In the last two sections, we respectively provide some concluding remarks for discussion and our future research lines.

\section{First-Order Versions of SUMO} \label{section:AdimenSUMO}

SUMO\footnote{\url{http://www.ontologyportal.org}} \cite{Niles+Pease'01} has its origins in the nineties, when a group of engineers from the IEEE Standard Upper Ontology Working Group pushed for a formal ontology standard. Their goal was to develop a standard upper ontology to promote data interoperability, information search and retrieval, automated inference and natural language processing.

SUMO is expressed in SUO-KIF (Standard Upper Ontology Knowledge Interchange Format \cite{Pea09}), which is a dialect of KIF (Knowledge Interchange Format \cite{Richard+'92}). Both KIF and SUO-KIF can be used to write FOL formulas, but its syntax goes beyond FOL. Consequently, SUMO cannot be directly used by FOL ATPs without a suitable transformation \cite{ALR12}.

There exist different proposals for converting large portions of SUMO into a FO ontology. In \cite{PeS07}, the authors report some preliminary experimental results evaluating the query timeout for different options when translating SUMO into FOL. Evolved versions of the translation described in \cite{PeS07} can be found in the TPTP Library\footnote{\url{http://www.tptp.org}} (hereinafter TPTP-SUMO). In \cite{ALR12}, we use ATPs for reengineering around 88\% of SUMO, obtaining Adimen-SUMO. Both TPTP-SUMO and Adimen-SUMO inherits information from the top and the middle levels of SUMO (from now on, the {\it core} of SUMO), thus discarding all the information  from the domain ontologies. In Table \ref{table:SUMOFigures}, we provide some figures comparing the explicit content of SUMO, TPTP-SUMO and Adimen-SUMO. It is easy to see that the most significant difference between TPTP-SUMO and Adimen-SUMO is the number of objects, which is due to the fact that TPTP-SUMO introduces many instances that should be inferred from the knowledge of the ontology.

An example of the practical inference capabilities of TPTP-SUMO and Adimen-SUMO is the CQ {\it ``Boys are not domestic animals''}
\begin{flalign}
\hspace{0.25cm} & ( \connective{=>} & \label{axiom:BoyDomesticAnimal} \\[-1mm]
 & \tab ( \predicate{instance} \; \variable{OBJ} \; \constant{Boy} ) & \nonumber \\[-1mm]
 & \tab ( \connective{not} & \nonumber \\[-1mm]
 & \tab \tab ( \predicate{instance} \; \variable{OBJ} \; \constant{DomesticAnimal} ) ) ) & \nonumber
\end{flalign}
that can be proved using ATPs.\footnote{In this paper, all axioms are to be considered universally closed.} In the next sections, we will also provide some CQs that cannot be inferred from neither TPTP-SUMO nor Adimen-SUMO.

\section{Competency Questions and First-Order Ontologies} \label{section:methodology}

In this section, we describe how to adapt the methodology introduced in \cite{GrF95} for evaluating and improving large and complex FO ontologies using state-of-the-art ATPs like Vampire \cite{RiV02} or E \cite{Sch02}.

In \cite{GrF95}, the authors propose to evaluate the expressiveness of an ontology by proving completeness theorems w.r.t. a set of CQs. The proof of completeness theorems requires to check whether a given CQ is entailed by the ontology or not. For this purpose, we use Vampire v3.0,\footnote{\url{http://www.vprover.org}} which works by refutation within a given execution-time limit. Similarly, we could also use E or other ATPs that work by refutation. Theoretically, if the conjecture is entailed by the ontology, then ATPs will eventually find a refutation given enough time (and space). However, theorem proving in FOL is a very hard problem, so it is not reasonable to expect ATPs to find a proof for every entailed conjecture \cite{KoV13}. Thus, if the ATP is able to find a prove for a conjecture, then we know for sure that the corresponding CQ is entailed by the ontology. However, if the ATP cannot find a proof, we do not know if (a) the conjecture is not entailed by the ontology or (b) although the conjecture is entailed, the ATP has not been able to find the proof within the provided execution-time limit. Due to the semi-decidability problem of FOL, increasing the execution-time limit is not a solution for conjectures that are not entailed. For the same reason, using other systems that do not work by refutation (for example, by model generation) is not a general solution. To overcome this problem, we consider three possibilities when testing the ontology w.r.t. a given CQ: the test may be (i) {\it passing}, (ii) {\it non-passing} or (iii) {\it unknown}, as we next describe.

As proposed in \cite{GrF95}, our method is based on a set of CQs written as conjectures in the language of the ontology. The set of CQs is partitioned into two classes: {\it truth-tests} and {\it falsity-tests}, depending on whether we expect the conjecture to be entailed by the ontology or not. For example, let us consider the CQs {\it ``Men cannot be pregnant''} and {\it ``Organisms cannot be dead''}:
\begin{flalign}
\hspace{0.25cm} & ( \connective{=>} & \label{axiom:ManPregnant} \\[-1mm]
 & \tab ( \connective{and} & \nonumber \\[-1mm]
 & \tab \tab ( \predicate{instance} \; \variable{HUMAN} \; \constant{Human} ) & \nonumber \\[-1mm]
 & \tab \tab ( \predicate{attribute} \; \variable{HUMAN} \; \constant{Pregnant} ) ) & \nonumber \\[-1mm]
 & \tab ( \connective{not} & \nonumber \\[-1mm]
 & \tab \tab ( \predicate{instance} \; \variable{HUMAN} \; \constant{Man} ) ) ) & \nonumber \\
 & ( \connective{=>} & \label{axiom:InmortalOrganisms} \\[-1mm]
 & \tab ( \predicate{instance} \; \variable{ORG} \; \constant{Organism} ) & \nonumber \\[-1mm]
 & \tab ( \connective{not} & \nonumber \\[-1mm]
 & \tab \tab ( \predicate{attribute} \; \variable{ORG} \; \constant{Dead} ) ) ) & \nonumber
\end{flalign}
According to the common sense knowledge, the first CQ is a truth-test, whereas the second one is a falsity-test (since {\it ``Organisms can be dead''}). Following this division, our method proceeds in two steps. In the first step, we deal with the set of truth-tests as conjectures. A truth-test is classified as {\it passing} if the ATP proves that the corresponding conjecture is entailed by the ontology, and it would be classified as {\it non-passing} if the ATP could prove that the conjecture is not entailed. However, as discussed above, when no proof is found we do not know whether the conjecture is entailed or not, thus we classify the truth-test as {\it unknown}. In the second step, we deal with the set of falsity-tests, which are supposed not to be inferred from the ontology. Hence, we classify a falsity-test as {\it non-passing} when the ATP proves that corresponding conjecture is entailed by the ontology, and as {\it unknown} when the ATP does not find any proof. 

In practice, unknown truth-tests are treated as non-passing, since ATPs have not been able to prove that the corresponding conjectures are entailed by the ontology. On the contrary, unknown falsity-tests are treated as passing. When a test is classified as non-passing (or unknown in the case of truth-tests), we proceed to correct the ontology. The correction is creative and directed by the problem at hand. In the case of truth-tests, we do not obtain any information for the tests that are not classified as {\it passing} (since ATPs find no proof), thus the correction is harder and even we do not know if any correction is required. With respect to falsity-tests, the proof generated by ATPs is used to isolate the controversial axioms, performing additional tests when necessary. The feedback provided by ATPs is very helpful for detecting modelling errors. As ATPs are continuously evolving, our method is producing more precise and useful outcomes, but the semi-decidability problem of FOL still remains. In frameworks where the underlying logic is decidable (like OWL-DL), our framework would be also applicable with the advantage of becoming exact.

In order to carry out the experiments described in this paper, we proceed as follows. First, we collect all creative queries available from the literature. In particular, the 33 questions from the CSR (Common Sense Reasoning) problem domain of TPTP (Thousands of Problems for Theorem Provers) \cite{Sut09} and the 5 questions described in \cite{ALR12}. Then, we extend this reduced set of 38 CQs with 26 new creative CQs. All these 64 creative CQs have been classified manually as 50 {\it truth-tests} (the 38 old plus 12 new CQs) and 14 {\it falsity-tests} (all new). When applying our framework, we use this reduced set of 64 creative CQs as a dataset for development (as explained in Section \ref{section:refining}). For improving the competency of the ontology, we only consider the results and traces returned by the ATPs when proving the CQs of this reduced set of creative tests. Second, for testing the competency of new FO versions of SUMO, we automatically create a very large set of CQs derived from WN and its mapping to SUMO (as described in Section \ref{section:benchmark}). In our framework, we use this large set of more than 7,000 automatic CQs as a dataset for testing (see Section \ref{section:evaluation}).

\section{Improving a First-Order Ontology} \label{section:refining}

In this section, we report on the experience of improving an FO ontology applying our framework and the set of 64 creative CQs. In particular, we provide some examples of truth- and falsity-tests, explaining how we have used them and the feedback provided by ATPs for improving our ontology. As a result of this process, we have derived Adimen-SUMO v2.4.

As discussed in the above section, we have to improve the ontology when the conjecture corresponding to a truth-test is not proven to be entailed, or when ATPs prove that the conjecture corresponding to a falsity-test is entailed. In the former case, we get no more feedback from ATPs, since no proof is found. Hence, we have to manually check the ontology to search for the modelling error. In the latter case, we obtain a proof, which includes the incorrect axioms. Roughly speaking, the modelling error can refer to ontological concepts (relations or objects) that are used to define the structure and main features of the ontology itself ---the so-called {\it basic} concepts--- or to concepts that serves to describe the knowledge that is contained in the ontology ---from now on, {\it non-basic} concepts---. For example,  \textPredicate{instance}, \textPredicate{subclass}, \textPredicate{disjoint}, \textPredicate{partition}, \textPredicate{attribute}, \textPredicate{subAttribute}, \textPredicate{contraryAttribute} and \textPredicate{exhaustiveAttribute} are basic concepts in Adimen-SUMO, whereas \textConstant{NullList} and \textConstant{Animal} are examples of non-basic concepts.  According to our experience, there are three typical types of errors than can be informally described as follows:
\begin{itemize}
\item Missing characterization: the modelling error refers to a non-basic concept \textPredicate{C}, but the ontology lacks the axiomatization of \textPredicate{C}. This is the simplest case, since the solution consists in axiomatizing \textPredicate{C}.
\item Too weak characterization: the modelling error refers to a non-basic concept \textPredicate{C} that is characterized, but in a way too weak. In this case, the solution consists in repairing the characterization (updating one or more axioms) of \textPredicate{C}.
\item Unsuitable characterization of basic concepts: the modelling error refers to a non-basic concept \textPredicate{C} which is well-characterized in terms of some basic concept \textPredicate{B}, but \textPredicate{B} is not suitable defined. This is the most complex case, since the solution may require modifying several basic definitions of the ontology, or even its structure.
\end{itemize}
However, the distinction between basic and non-basic concepts in an ontology is not always clearly stated and is usually dependent upon one's interpretation. This is the case of SUMO, where all concepts are defined in terms of SUMO itself, preventing a direct translation of SUMO into FOL \cite{ALR12}. This problem also prevents a more formal characterization of modelling errors. Next, we provide some examples of the modelling errors described above and the CQs that have enabled its automated detection by following our proposal.

Regarding truth-tests, let us consider the CQs {\it ``A list containing at least one item is not empty (null)''} (\ref{axiom:NonNullList}), {\it ``Tables do not have a brain''} (\ref{axiom:BrainTablePart}) and {\it ``Men cannot be pregnant''} (\ref{axiom:ManPregnant}):
\begin{flalign}
\hspace{0.25cm} & ( \connective{=>} & \label{axiom:NonNullList} \\[-1mm]
 & \tab ( \connective{and} & \nonumber \\[-1mm]
 & \tab \tab ( \predicate{instance} \; \variable{LIST} \; \constant{List} ) & \nonumber \\[-1mm]
 & \tab \tab ( \predicate{instance} \; \variable{ITEM} \; \constant{Entity} ) & \nonumber \\[-1mm]
 & \tab \tab ( \predicate{inList} \; \variable{ITEM} \; \variable{LIST} ) ) & \nonumber \\[-1mm]
 & \tab ( \connective{not} \nonumber \\[-1mm]
 & \tab \tab ( \predicate{equal} \; \variable{LIST} \; \constant{NullList} ) ) ) & \nonumber \\
\hspace{0.25cm} & ( \connective{=>} & \label{axiom:BrainTablePart} \\[-1mm]
 & \tab ( \connective{and} & \nonumber \\[-1mm]
 & \tab \tab ( \predicate{instance} \; \variable{BRAIN} \; \constant{Brain} ) & \nonumber \\[-1mm]
 & \tab \tab ( \predicate{instance} \; \variable{TABLE} \; \constant{Table} ) ) & \nonumber \\[-1mm]
 & \tab ( \connective{not} & \nonumber \\[-1mm]
 & \tab \tab ( \predicate{properPart} \; \variable{BRAIN} \; \variable{TABLE} ) ) ) & \nonumber
\end{flalign}
After some initial experiments with ATPs, we realize that these conjectures are entailed from neither Adimen-SUMO nor TPTP-SUMO. Analysing the knowledge required to answer each question, we find the following problems.

In the case of the CQ {\it ``A list containing at least one item is not empty (null)''}, the object \textConstant{NullList} is only axiomatized to be instance of \textConstant{List}, hence a proper characterization is missing. To solve this problem, we include the following axiom in Adimen-SUMO v2.4 as characterization of \textConstant{NullList}:
\begin{flalign} \label{axiom:NullList}
\hspace{0.25cm} & ( \connective{not} & \\[-1mm]
 & \tab ( \predicate{inList} \; \variable{ITEM} \; \constant{NullList} ) ) & \nonumber
\end{flalign}
Including this axiom, Adimen-SUMO v2.4 entails {\it ``A list containing at least one item is not empty (null)''} (\ref{axiom:NonNullList}).

The second CQ {\it ``Tables do not have a brain''} (\ref{axiom:BrainTablePart}) cannot be proved because a too weak characterization of the concept \textConstant{Animal}. In particular, the source of the problem is:
\begin{flalign} \label{axiom:OldAnimalAnatomicalStructure}
\hspace{0.25cm} & ( \connective{=>} & \\[-1mm]
 & \tab ( \connective{and} & \nonumber \\[-1mm]
 & \tab \tab ( \predicate{instance} \; \variable{STRUCTURE} \; \constant{AnimalAnatomicalStructure} ) & \nonumber \\[-1mm]
 & \tab \tab ( \predicate{instance} \; \variable{ANIMAL} \; \constant{Organism} ) & \nonumber \\[-1mm]
 & \tab \tab ( \predicate{part} \; \variable{STRUCTURE} \; \variable{ANIMAL} ) ) & \nonumber \\[-1mm]
 & \tab ( \predicate{instance} \; \variable{ANIMAL} \; \constant{Animal} ) ) & \nonumber
\end{flalign}
This axiom can only be applied to instances of \textConstant{Organism}, which is not the case of \textConstant{Table}. However, it is not necessary to restrict the use of this axiom to instances of \textConstant{Organism}, since the ontology already entails that only \textConstant{Animal}s can have \textConstant{AnimalAnatomicalStructure}s. Thus, we could relax the antecedent of the formula by simply removing that restriction. However, \textPredicate{part} is defined as \textConstant{PartialOrderingRelation}, hence it is reflexive, and the classes \textConstant{AnimalAnatomicalStructure} and \textConstant{Animal} are defined to be disjoint. Thus, from the resulting axiom it would be possible to infer that any instance of \textConstant{AnimalAnatomicalStructure}, which is trivially part of itself, is also instance of \textConstant{Animal}, contradicting the disjointness of these classes. Therefore, it is more suitable the use of \textPredicate{properPart} (which is irreflexive) in the resulting axiom instead of \textPredicate{part}. To sum up, the above axiom (\ref{axiom:OldAnimalAnatomicalStructure}) is rewritten to:
\begin{flalign} \label{axiom:NewAnimalAnatomicalStructureProperpart}
\hspace{0.25cm} & ( \connective{=>} & \\[-1mm]
 & \tab ( \connective{and} & \nonumber \\[-1mm]
 & \tab \tab ( \predicate{instance} \; \variable{STRUCTURE} \; \constant{AnimalAnatomicalStructure} ) & \nonumber \\[-1mm]
 & \tab \tab ( \predicate{properPart} \; \variable{STRUCTURE} \; \variable{ANIMAL} ) ) & \nonumber \\[-1mm]
 & \tab ( \predicate{instance} \; \variable{ANIMAL} \; \constant{Animal} ) ) & \nonumber
\end{flalign}
Adimen-SUMO v2.4 entails {\it ``Tables do not have a brain''} by means of replacing the old axiom (\ref{axiom:OldAnimalAnatomicalStructure}) with (\ref{axiom:NewAnimalAnatomicalStructureProperpart}).

The problem of non-passing the truth-test {\it ``Men cannot be pregnant''} (\ref{axiom:ManPregnant}) comes from an unsuitable characterization of some basic relations about attributes. More specifically, \textPredicate{contraryAttribute} and \textPredicate{exhaustiveAttribute} are variable arity relations that constrain the use of attributes. As in the case of the other variable arity relations defined in SUMO (such as \textPredicate{partition}, \textPredicate{disjointDecomposition} or \textPredicate{exhaustiveDecomposition}), \textPredicate{contraryAttribute} and \textPredicate{exhaustiveAttribute} are characterized on the basis of \textPredicate{inList} and \textPredicate{ListOrderFn}. In Adimen-SUMO, the so-called {\it row operators} are used for a proper characterization \cite{ALR12} (this problem remains unsolved in TPTP-SUMO). However, Adimen-SUMO v2.2 directly inherited from SUMO unsuitable characterizations of \textPredicate{inList} and \textPredicate{ListOrderFn} that prevent to prove most of the truth-tests involving attributes. To fix this problem, we have included in Adimen-SUMO v2.4 two axioms characterizing \textPredicate{contraryAttribute} and \textPredicate{exhaustiveAttribute}, which enables Adimen-SUMO v2.4 to entail {\it ``Men cannot be pregnant''}.

Now, let us consider the falsity-tests {\it ``Tables can be living''} (\ref{axiom:TableLiving}) and {\it ``Organisms cannot be dead''} (\ref{axiom:InmortalOrganisms}):
\begin{flalign} \label{axiom:TableLiving}
\hspace{0.25cm} & ( \connective{exists} ( \variable{TABLE} ) \\[-1mm]
 & \tab ( \connective{and} & \nonumber \\[-1mm]
 & \tab \tab ( \predicate{instance} \; \variable{TABLE} \; \constant{Table} ) & \nonumber \\[-1mm]
 & \tab \tab ( \predicate{attribute} \; \variable{TABLE} \; \constant{Living} ) ) ) & \nonumber
\end{flalign}
We have performed many runs using different ATPs and none of them finds a proof of the goals (\ref{axiom:TableLiving}) and (\ref{axiom:InmortalOrganisms}) from Adimen-SUMO v2.2 or TPTP-SUMO. Likewise, ATPs do not find either any refutation for (\ref{axiom:TableLiving}) from Adimen-SUMO v2.4. However, the new characterization of \textPredicate{contraryAttribute} and \textPredicate{exhaustiveAttribute} in Adimen-SUMO v2.4 enables ATPs to find a proof of {\it ``Organisms cannot be dead''} (\ref{axiom:InmortalOrganisms}). From this proof, we discover that the problem is related with the following axioms:
\begin{flalign}
\label{axiom:DeadLivingContraryAttribute}
\hspace{0.25cm}  & ( \predicate{contraryAttribute} \; \constant{Dead} \; \constant{Living} ) & \\
\label{axiom:DeadUnconscious}
 & ( \predicate{subAttribute} \; \constant{Dead} \; \constant{Unconscious} ) & \\
\label{axiom:UnconsciousConsciousnessAttribute}
 & ( \predicate{instance} \; \constant{Unconscious} \; \constant{ConsciousnessAttribute} ) \hspace{-1cm} & \\
\label{axiom:ConsciousnessAttributeLiving}
 & ( \connective{<=>} & \\[-1mm]
 & \tab ( \connective{and} & \nonumber \\[-1mm]
 & \tab \tab ( \predicate{instance} \; \variable{AGENT} \; \constant{SentientAgent} ) & \nonumber \\[-1mm]
 & \tab \tab ( \predicate{attribute} \; \variable{AGENT} \; \constant{Living} ) ) & \nonumber \\[-1mm]
 & \tab ( \connective{exists} \; ( \variable{ATTR} ) & \nonumber \\[-1mm]
 & \tab \tab ( \connective{and} & \nonumber \\[-1mm]
 & \tab \tab \tab ( \predicate{instance} \; \variable{ATTR} \; \constant{ConsciousnessAttribute} ) & \nonumber \\[-1mm]
 & \tab \tab \tab ( \predicate{attribute} \; \variable{AGENT} \; \variable{ATTR} ) ) ) ) & \nonumber
\end{flalign}
According to axiom (\ref{axiom:DeadLivingContraryAttribute}), it is not possible to have both \textConstant{Living} and \textConstant{Dead} as attribute. However, by (\ref{axiom:DeadUnconscious}) and (\ref{axiom:UnconsciousConsciousnessAttribute}), \textConstant{Dead} is an instance of \textConstant{ConciousnessAttribute}. Moreover, having \textConstant{Dead} as attribute implies to also have \textConstant{Living} by (\ref{axiom:ConsciousnessAttributeLiving}). Analysing this set of conflictive axioms, we have decided to remove axiom (\ref{axiom:DeadUnconscious}), which defines \textConstant{Dead} as subattribute of \textConstant{Unconscious}, from Adimen-SUMO v2.4. This incorrectness was hidden in both Adimen-SUMO 2.2 and TPTP-SUMO due to the inappropriate characterization of attributes.

After this development phase, Adimen-SUMO v2.4 passes the 50 creative truth-tests, whereas 23 and 15 creative truth-tests are classified as unknown by TPTP-SUMO and Adimen-SUMO v2.2 respectively. Regarding the creative falsity-tests, all of them are classified as unknown by the three ontologies.

\section{Deriving Competency Questions from WordNet} \label{section:benchmark}

In order to build our benchmark, we have used the mapping from WN to SUMO \cite{Niles+Pease'03}. This mapping connects each synset of WN into a term of SUMO using three relations: {\it equivalence}, {\it subsumption} and {\it instance}. These relations will be denoted by concatenating the symbols `=' ({\it equivalence}), `+' ({\it subsumption}) and `@' ({\it instance}) to the corresponding SUMO concept. For example, {\it piloting$_n^2$}, {\it education$_n^4$} and {\it zero$_a^1$} are connected to \textConstant{Pilot}=, \textConstant{EducationalProcess}+ and \textConstant{Integer}@. Additionally, the complementary of the relations {\it equivalence} and {\it subsumption} are also used.

The mapping from WN to SUMO uses terms from the core of SUMO, but also from the domain ontologies. However, both TPTP-SUMO and Adimen-SUMO only use axioms from the core of SUMO. Thus, our first task has been to obtain a mapping from WN to the core of SUMO on the basis of the mapping from WN to SUMO. To this end, for each WN synset not mapped to a term covered by both TPTP-SUMO and Adimen-SUMO, we have conveniently used the structural relations of SUMO (\textPredicate{instance}, \textPredicate{subclass}, \textPredicate{subrelation} and \textPredicate{subAttribute}) to inherit the term of the core of SUMO to which the synset is connected. Note that this process sometimes requires to modify the mapping relation. For example, the synset {\it frying$_n^1$} is connected to the SUMO class \textConstant{Frying}=, which belongs to the domain ontology {\it Food}. In the same domain ontology, \textConstant{Frying} is defined to be subclass of \textConstant{Cooking}, which is defined in the top level of SUMO. Consequently, \textConstant{Frying} is not defined in the core of SUMO, but \textConstant{Cooking} is. Thus, the synset {\it frying$_n^1$} can be connected to \textConstant{Cooking} in the resulting mapping. However, instead of {\it equivalence}, {\it frying$_n^1$} is connected to \textConstant{Cooking} by the {\it subsumption} mapping relation: that is, \textConstant{Cooking}+. The total number of mappings to the core of SUMO (114,948) is slightly smaller than the number of mappings to SUMO (115,872) since some terms are not properly defined. This is mainly due to the fact that some terms in the mapping derived from older versions of SUMO are not longer available in the current one. For example, the synsets {\it salmon$_n^1$} and {\it architect$_n^2$} are respectively connected to the SUMO concepts \textConstant{Salmon}= and \textConstant{Architect}=, which do not appear in the latest versions of SUMO.

After obtained a suitable mapping from WN to the ontologies that we want to compare, we have designed several conceptual patterns of questions regarding the information about antonyms and processes in WN. 

\subsection{Antonym patterns}

WN provides a set of 8,689 antonym-pairs, including nouns, verbs, adjectives and adverbs, from which 7,410 antonym-pairs can be properly mapped to the core of SUMO, as described above. However, we only consider the antonym-pairs where both synsets are connected using the {\it equivalence} mapping relation (in total, 190 pairs), discarding those where the {\it subsumption} mapping relation is used. For these 190 antonym-pairs, we propose two conceptual patterns of questions. The first pattern is based on the fact that two SUMO classes connected to antonym synsets of WN cannot have common instances. For example, the antonym synsets {\it frozen$_n^1$} and {\it liquescent$_n^1$} are respectively connected to \textConstant{Freezing}= and \textConstant{Melting}=. Thus, from the above antonym-pair, we derive the following competency question:
\begin{flalign} \label{CQ:MeltingFreezing}
\hspace{0.25cm} & ( \connective{not} & \\[-1mm]
 & \tab ( \connective{exists} ( \variable{X} ) & \nonumber \\[-1mm]
 & \tab \tab ( \connective{and} & \nonumber \\[-1mm]
 & \tab \tab \tab ( \predicate{instance} \; \variable{X} \; \constant{Melting} ) & \nonumber \\[-1mm]
 & \tab \tab \tab ( \predicate{instance} \; \variable{X} \; \constant{Freezing} ) ) ) ) & \nonumber
\end{flalign}
Similarly, the second conceptual pattern states that two attributes connected to antonym synsets are not compatible. For example, from the antonym synsets {\it waking$_n^1$} and {\it sleeping$_n^1$}, which are connected to \textConstant{Awake}= and \textConstant{Asleep}=, we derive the following competency question:
\begin{flalign} \label{CQ:AwakeAsleep}
\hspace{0.25cm} & ( \connective{not} & \\[-1mm]
 & \tab ( \connective{exists} ( \variable{X} ) & \nonumber \\[-1mm]
 & \tab \tab ( \connective{and} & \nonumber \\[-1mm]
 & \tab \tab \tab ( \predicate{attribute} \; \variable{X} \; \constant{Awake} ) & \nonumber \\[-1mm]
 & \tab \tab \tab ( \predicate{attribute} \; \variable{X} \; \constant{Asleep} ) ) ) ) & \nonumber
\end{flalign}
Applying these two patterns on the 190 antonym-pairs where both synsets are connected used {\it equivalence}, we obtain 64 different truth-tests. By negating each of the above 64 CQs, we also obtain 64 different falsity-tests.

\begin{table*}
\caption{\label{table:Comparison} Evaluation of SUMO-based FO ontologies} 
\centering
\resizebox{\textwidth}{!}{
\begin{tabular}{lrrrrr|rrrrr|rrrrr}
\hline \\[-6pt]
\multicolumn{1}{c}{Tests} & \multicolumn{5}{c}{TPTP-SUMO} & \multicolumn{5}{c}{Adimen-SUMO v2.2} & \multicolumn{5}{c}{Adimen-SUMO v2.4} \\
\hline \\[-6pt]
 & \multicolumn{1}{c}{P} & \multicolumn{1}{c}{N} & \multicolumn{1}{c}{U} & \multicolumn{1}{c}{t(P)} & \multicolumn{1}{c}{t(N)} & \multicolumn{1}{c}{P} & \multicolumn{1}{c}{N} & \multicolumn{1}{c}{U} & \multicolumn{1}{c}{t(P)} & \multicolumn{1}{c}{t(N)} & \multicolumn{1}{c}{P} & \multicolumn{1}{c}{N} & \multicolumn{1}{c}{U} & \multicolumn{1}{c}{t(P)} & \multicolumn{1}{c}{t(N)} \\
\hline \\[-6pt]
Truth-tests (3,556) & 4 & -- & 3,552 & 361.97 s. & --~ & 89 & -- & 3,467 & 56.06 s. & --~ & 894 & -- &  2,662 & 56.46 s. & --~ \\
\hspace{0.25cm} Antonym pattern (64) & 3 & -- & 61 & 473.26 s. & --~ & 17 & -- & 47 & 6.77 s. & --~ & 45 & -- & 19 & 32.18 s. & --~ \\
\hspace{0.25cm} Relation pattern (1,280) & 0 & -- & 1,280 & --~ & --~ & 11 & -- & 1,269 & 121.09 s. & --~ & 176 & -- & 1,104 & 63.72 s. & --~ \\
\hspace{0.25cm} Event pattern \#1 (25) & 0 & -- & 25 & --~ & --~ & 2 & -- & 23 & 18.72 s. & --~ & 7 & -- & 18 & 108.30 s. & --~ \\
\hspace{0.25cm} Event pattern \#2 (330) & 0 & -- & 330 & --~ & --~ & 26 & -- & 304 & 45.46 s. & --~ & 115 & -- & 215 & 44.16 s. & --~ \\
\hspace{0.25cm} Event pattern \#3 (1,857) & 1 & -- & 1,756 & 28.13 s. & --~ & 33 & -- & 1,824 & 70.40 s. & --~ & 551 & ~-- & 1,306 & 58.03 s. & --~ \\
Falsity-tests (3,556) & -- & 466 & 3,090 & --~ & 25.19 s. & -- & 493 & 3,063 & --~ & 6.89 s. & -- & 487 & 3,069 & --~ & 6.88 s. \\
\hspace{0.25cm} Antonym pattern (64) & -- & 4 & 60 & --~ & 22.16 s. & -- & 2 & 62 & --~ & 3.31 s. & -- & 5 & 59 & --~ & 14.06 s. \\
\hspace{0.25cm} Relation pattern (1,280) & -- & 4 & 1,276 & --~ & 191.93 s. & -- &  31 & 1,249 & --~ & 97.69 s. & -- & 22 & 1,258 & --~ & 114.60 s. \\
\hspace{0.25cm} Event pattern \#1 (25) & -- & 0 & 25 & --~ & --~ & -- & 0 & 25 & --~ & --~ & -- & 0 & 25 & --~ & --~ \\
\hspace{0.25cm} Event pattern \#2 (330) & -- & 71 & 259 & --~ & 23.73 s.  & -- & 72 & 258 & --~ & 0.57 s. & -- & 72 & 258 & --~ & 1.18 s. \\
\hspace{0.25cm} Event pattern \#3 (1,857) & -- & 387 & 1,470 & --~ &  23.76 s. & -- & 388 & 1,469 & --~ & 0.82 s. & -- & 388 & 1,469 & --~ & 1.73 s. \\
\multicolumn{16}{c}{~}\\[-0.3cm]
\hline
\end{tabular}
}
\end{table*}

\subsection{Process patterns}

Regarding processes, we have used the information in the morphosemantic database,\footnote{Available at \url{http://wordnetcode.princeton.edu/standoff-files/morphosemantic-links.xls}.} which contains semantic relations between morphologically related nouns and verbs. From the 14 semantic relations defined in the database, we select {\it agent}, {\it result}, {\it instrument} and {\it event}. The first three ones relate a process (verb) which its corresponding agent / result / instrument (noun), from which we infer 1,280 CQs by simply stating the same property in terms of SUMO. For example, WN establishes that the {\it result} of {\it compose$_v^2$} is a {\it composition$_n^4$}, which are respectively mapped to \textConstant{ComposingMusic}+ and \textConstant{MusicalComposition}=:
\begin{flalign} \label{axiom:ResultComposingMusic}
\hspace{0.25cm} & ( \connective{exists} ( \variable{X} \; \variable{Y} ) \\[-1mm]
 & \tab ( \connective{and} & \nonumber \\[-1mm]
 & \tab \tab ( \predicate{instance} \; \variable{X} \; \constant{ComposingMusic} ) & \nonumber \\[-1mm]
 & \tab \tab ( \predicate{result} \; \variable{X} \; \variable{Y} ) & \nonumber \\[-1mm]
 & \tab \tab ( \predicate{instance} \; \variable{Y} \; \constant{MusicalComposition} ) ) ) & \nonumber
\end{flalign}
As before, we also obtain 1,280 falsity-tests by negating the previous ones. 

The last relation {\it event} connects nouns and verbs referring to the same process. Being the same process, we assume that both the noun and the verb should be mapped to the same class of SUMO. Thus, if the noun and the verb are mapped to different SUMO class constants, our hypothesis is that the mapping is wrong. Following this criterion, we propose different conceptual patterns of questions, depending on the used mapping relations, with the purpose of detecting wrong-mappings. If both synsets are connected to two different SUMO class constants using the {\it equivalence} mapping relation, it should be possible to prove that the two class constants denote different classes. For example, {\it kill$_v^{10}$} and {\it killing$_n^2$} are respectively connected to the SUMO classes \textConstant{Death}= and \textConstant{Killing}=, hence we derive the following CQ:
\begin{flalign} \label{CQ:DeathKilling}
\hspace{0.25cm} & ( \connective{not} \\[-1mm]
 & \tab ( \predicate{equal} \; \constant{Death} \; \constant{Killing} ) ) & \nonumber
\end{flalign}
Using this pattern, we derive 25 CQs. The second pattern of question focuses on the case where the synsets are connected to different SUMO class constants and using different mapping relations. That is, one synset is connected using the {\it equivalence} mapping relation, whereas the other synset is connected using {\it subsumption}. Being the mapping information less precise than in the first case, it does not suffice to prove that the classes are different. In this case, the pattern states that the class connected using {\it equivalence} cannot be subclass of the class connected using {\it subsumption}. For example, {\it event} relates {\it repair$_n^1$}, which is connected to \textConstant{Repairing}=, and {\it repair$_v^1$}, which is connected to \textConstant{Pretending}+. Therefore, we assume that \textConstant{Repairing} cannot be subclass of \textConstant{Pretending}, deriving the following CQ:
\begin{flalign} \label{CQ:RepairingPretending}
\hspace{0.25cm} & ( \connective{not} \\[-1mm]
 & \tab ( \predicate{subclass} \; \constant{Repairing} \; \constant{Pretending} ) ) & \nonumber
\end{flalign}
From the second {\it event} pattern of questions, we derive 330 CQs. In fact, this second pattern can be seen as a particular case of the third one, where both synsets are connected using the {\it subsumption} mapping relation. In this case, the pattern states that none of the connected SUMO classes can be subclass of the other one. For example, {\it event} relates the synsets {\it measure$_v^4$} and {\it appraisal$_n^1$}, which are respectively connected to \textConstant{Judging}+ and \textConstant{Comparing}+. Consequently, we derive the following CQ:
\begin{flalign} \label{CQ:JudgingComparing}
\hspace{0.25cm} & ( \connective{not} & \\[-1mm]
 & \tab  ( \connective{or} & \nonumber\\[-1mm]
 & \tab \tab ( \predicate{subclass} \; \constant{Judging} \; \constant{Comparing} ) & \nonumber \\[-1mm]
 & \tab \tab ( \predicate{subclass} \; \constant{Comparing} \; \constant{Judging} ) ) ) & \nonumber
\end{flalign}
Using this third pattern, we obtain 1,857 CQs. 

In total, we obtain 2,212 truth-tests by stating that the mapping is not correct, and the corresponding 2,212 falsity-tests stating that the mapping is correct.

\section{Evaluating First-order Ontologies} \label{section:evaluation}

In this section, we summarize the evaluation results of TPTP-SUMO and the different versions of Adimen-SUMO using the methodology proposed in Section \ref{section:methodology}. For this evaluation, we have used the set of 7,112 CQs that have been automatically obtained from WN, as described in the above section.

Table \ref{table:Comparison} sums up some runtime figures of the ATP Vampire 3.0\footnote{\url{http://www.vprover.org}} \cite{RiV02} when evaluating TPTP-SUMO and Adimen-SUMO with an execution time limit of 600 seconds.\footnote{In this experimentation, we have used a standard 64-bit Intel\textregistered~Core\texttrademark~i7-2600 CPU @ 3.40GHz desktop machine with 16GB of RAM.} For each ontology, we provide the number of passing ({\it P} column), non-passing ({\it N} column) and unknown CQs ({\it U} column), together with the average runtimes of passing ({\it t(P)} column) and non-passing ({\it t(N)} column) CQs. It is worth to remark that the average runtime of the CQs classified as unknown is the maximum execution time (600 seconds), since no proof is found. From the results, it is clear that Adimen-SUMO v2.4 outperforms Adimen-SUMO v2.2 in terms of competency in both the truth-test (more passing tests) and the falsity-test category (less non-passing tests). Regarding efficiency, the average runtime of Adimen-SUMO v2.4 is longer since the passed additional tests require more complex proofs. Similarly, Adimen-SUMO v2.2 outperforms TPTP-SUMO in the truth-test category,  since TPTP-SUMO only passes 4 truth-tests while Adimen-SUMO v2.2 passes 89. Regarding falsity-tests, TPTP-SUMO is the ontology with less non-passing tests, but the average runtime is clearly longer. Thus, we think that the number of non-passing falsity-tests of TPTP-SUMO would be larger if we used a longer execution time limit for the experimentation.

Recall that non-passing falsity-tests can provide useful information to improve the ontology, as in the case of the CQ {\it ``Organisms cannot be dead''} (see Section \ref{section:refining}). Additionally, non-passing falsity-tests also provide very useful information. For example, the verbs {\it whisper$_v^1$} and {\it shout$_v^1$}, which are antonyms in WN, are mapped to the SUMO classes \textConstant{Speaking}= and \textConstant{Vocalizing}= respectively. Being antonyms, we expect that these two SUMO classes, \textConstant{Speaking} and \textConstant{Vocalizing}, do not have any common instance, as stated by the next conjecture which corresponds to a CQ included in the automatic falsity-test set:
\begin{flalign} \label{axiom:AntonymVocalizingSpeaking}
\hspace{0.25cm} & ( \connective{exists} ( \variable{X} ) \\[-1mm]
 & \tab ( \connective{and} & \nonumber \\[-1mm]
 & \tab \tab ( \predicate{instance} \; \variable{X} \; \constant{Vocalizing} ) & \nonumber \\[-1mm]
 & \tab \tab ( \predicate{instance} \; \variable{X} \; \constant{Speaking} ) ) ) & \nonumber
\end{flalign}
However, ATPs can infer the above conjecture from Adimen-SUMO v2.4. That is, Adimen-SUMO v2.4 does not pass this falsity-test. This fact serves to detect that the mapping of the verbs {\it whisper$_v^1$} and {\it shout$_v^1$} to SUMO is not suitable. On the contrary, the CQ is classified as unknown when evaluating TPTP-SUMO and Adimen-SUMO v2.2, which prevents to detect the incorrect mapping.

\section{Discussion} \label{section:discussion}

We have shown a new framework and experimental results for evaluating and improving large and complex FO ontologies using ATPs. Our results show the appropriateness of using ATPs in debugging FO ontologies, as well as their practical use for inferring non-trivial statements. 

Although the authors of \cite{BeZ13} claim for the necessity of proving the formal faithfulness of a logical translation of SUMO, the formats KIF and SUO-KIF lack a formal notion of logical consequence and a formal deduction system,\footnote{A formal declarative semantics for the FO sublanguage SKIF of KIF is given in \cite{Hayes+Menzel'01}.} where that kind of mathematical results (such as conservative extension) relies on. Thus, following \cite{ALR12}, our repairs just intend to improve the reasoning capabilities of the ontology. 

Another possible topic of discussion is the need for some clear {\it quality criteria} of the CQs. For instance, all our CQs are universally closed formulas of the form
\begin{equation} \label{eq:goal}
( P_1 \wedge P_2 \wedge \ldots \wedge P_n ) \to P
\end{equation}
where $P$ is not deducible from $\{P_1,P_2, \dots, P_n\}$ without the help of the knowledge contained in the ontology. That is, none of our CQs includes information in the premises that should be inferred from the ontology. Note that this is not the case of some of the tests from the CSR problem domain of TPTP \cite{Sut09}. Thus, our set of CQs includes a {\em clean} version of these goals all the properties $P_i$ in (\ref{eq:goal}) that should be deducible from the ontology have been removed. 

Finally, our set of CQs do not only have conjectures that are expected to be deducible, but it also contains conjectures expected not to be deducible. Obviously, this is also a very interesting experimentation, since these conjectures allow the detection of modelling errors when proved.

\section{Future Work} \label{section:futurework}

The new framework presented in this paper opens multiple avenues for future research. Now, we are working in order to improve the competency of Adimen-SUMO. In the case of non-passing automatic tests, this implies to correct either a) the ontology itself, b) some mappings from WN to the ontology, or c) some WN relations. In parallel, we also want to enlarge our current set of CQs by gathering more questions from WN and its mapping to SUMO or alternative datasets. Additionally, it would be very interesting to determine which parts of the ontology are used to solve a set of CQs. Thus, in order to further debug Adimen-SUMO, we are also exploring the possibility to automatically derive an exhaustive set of questions from its axiomatization.

Our framework does not only allow to measure the competency of different SUMO-based ontologies, but also its efficiency when solving a large set of non-trivial inferences. In this sense, we are investigating more efficient representations of the ontology. Additionally, our framework can act as a new benchmark for testing the performance of FOL ATPs.

Another open research line will focus on proving the consistency of Adimen-SUMO. For instance, on the basis of a modular approach \cite{KuM11} or by including into our framework sophisticated model finders of the type of iProver \cite{Kor08}.

Finally, we plan to develop automatic procedures to exploit Adimen-SUMO and its complete mapping to WN for automatically inferring new semantic properties and relations between WN concepts, or validating the consistency of resources associated to WN such as Cyc, DBpedia or Yago.

\section{Acknowledgments}

We are grateful to the anonymous reviewers for their insightful comments. This work has been partially funded by the Spanish Projects SKaTer (TIN2012-38584-C06-02) and COMMAS\; (TIN2013-46181-C2-2-R),\; the\; Basque\; Project LoRea (GIU12/26) and grant BAILab (UFI11/45).

\bibliographystyle{abbrv}
\bibliography{alr15}

\end{document}